\documentclass{svproc}
\usepackage{graphicx}
\usepackage{amsmath}
\usepackage{hyperref}

\begin{document}
\mainmatter              
\title{Energy Efficient Hardware Acceleration of Neural Networks with Power-of-Two Quantisation}
\titlerunning{Hardware Acceleration of NNs with PoT Quantisation}  
%
\author{Dominika Przewlocka-Rus \href{https://orcid.org/0000-0002-5836-8604}{\includegraphics[width=16pt]{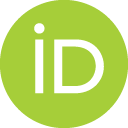}} 
   \and Tomasz Kryjak \href{https://orcid.org/0000-0001-6798-4444}{\includegraphics[width=16pt]{orcid.png}}}
\authorrunning{D. Przewlocka-Rus, T. Kryjak} 
%
%
\institute{Embedded Vision Systems Group, Computer Vision Laboratory, \\ Department of Automatic Control and Robotics, \\ AGH University of Science and Technology, Krakow, Poland\\
\email{\{dominika.przewlocka,tomasz.kryjak\}@agh.edu.pl}}

\maketitle              

\begin{abstract}

Deep neural networks virtually dominate the domain of most modern vision systems, providing high performance at a cost of increased computational complexity. 
Since for those systems it is often required to operate both in real-time and with minimal energy consumption (e.g., for wearable devices or autonomous vehicles, edge Internet of Things (IoT), sensor networks), various network optimisation techniques are used, e.g., quantisation, pruning, or dedicated lightweight architectures. 
Due to the logarithmic distribution of weights in neural network layers, a method providing high performance with significant reduction in computational precision (for 4-bit weights and less) is the Power-of-Two (PoT) quantisation (and therefore also with a logarithmic distribution). This method introduces additional possibilities of replacing the typical for neural networks Multiply and ACcumulate (MAC -- performing, e.g., convolution operations) units, with more energy-efficient Bitshift and ACcumulate (BAC). 
In this paper, we show that a hardware neural network accelerator with PoT weights implemented on the Zynq UltraScale + MPSoC ZCU104 SoC FPGA can be at least $1.4x$ more energy efficient than the uniform quantisation version. 
To further reduce the actual power requirement by omitting part of the computation for zero weights, we also propose a new pruning method adapted to logarithmic quantisation.

\end{abstract}
\section{Introduction}
Computer vision algorithms are a central component in many of the most modern and emerging technologies, such as autonomous vehicles (drones, cars), augmented and virtual reality, and advanced driver assistance systems (ADAS).
These systems must be characterised with three features: reliability, real-time operation, and, due to the battery power source, energy efficiency.
The highest performance of vision systems is currently guaranteed by machine learning algorithms, in particular deep neural networks, which at the same time are known for high computational complexity.
To guarantee real-time operation, these calculations need to be massively parallelised, with a third factor -- the requirement for energy efficiency -- to be taken into account.
Since regular CPUs are not designed for such multi-threaded operations, and high-end GPUs consume hundreds of watts, embedded devices seem to be a~promissing solution.
However, these platforms are relatively small in computational resources and neural network acceleration requires some model optimisation.
The most commonly used methods to reduce the number of parameters, as well as the computational complexity of the network, are quantisation and pruning.

\begin{figure}[!t]
\begin{center}
\includegraphics[width=\textwidth]{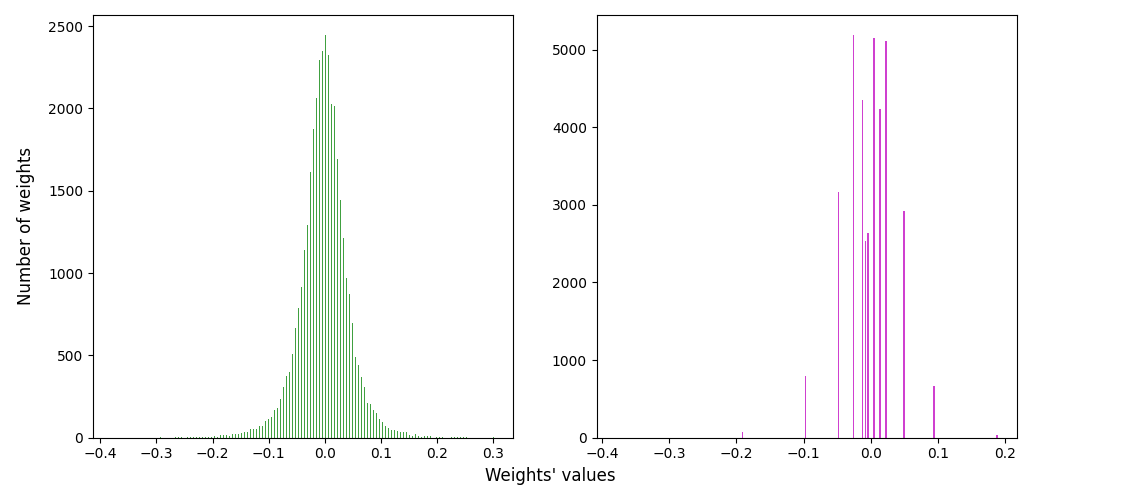}
\caption{The left plot shows the distribution of an exemplar layer of ResNet18 trained on ImageNet, whose logarithmic shape motivates the use of the Power-of-Two quantisation method. The weights' distribution after quantisation to the 4 bit width is shown on the right.} \label{fig::pot_quantization}
\end{center}
\end{figure}

Reducing the precision of a computation using quantisation leads to a significant reduction in memory requirements -- e.g., going from 32-bit floating-point numbers to an 8-bit integer representation, the needed storage is reduced $4x$ --, but also a simplification of the computations is achieved (multiplying integers is less complex than multiplying floating-point numbers).
The gain will be greater with a significant reduction in precision, such as to 4 or even fewer bits.
Not to lead to a clear decrease in network performance, one should use a quantisation method which takes into account the logarithmic distribution of weights in the neural network layer, which are concentrated around zero and more sparse elsewhere.
One possibility is to use weights that are powers of two (PoT weights) -- Fig. \ref{fig::pot_quantization}.
Such quantisation has another major advantage, as it allows multiplication operation (in convolution filters and fully connected layers) to be converted into a very simple bit-shift operation.
Another method reducing the memory complexity of the network is pruning, which zeroes very small weights, as they have little effect on the final result of input processing by the network.


In this paper, we focus on demonstrating the energy gains for FPGA-based (Filed Programmable Gate Array) neural network accelerators with PoT weights.
FPGA devices are programmed by defining connections between the electronic components available in the device. 
They allow massive parallelisation of the operations and usually consume only a few watts.
With the appropriate methods of model optimisation, FPGAs can ensure the fulfilment of all the constraints discussed for the vision systems.

The main contributions of this paper are:
\begin{itemize}
    \item Design of a Bitshift ACcumulate (BAC) module implementing convolutional filtering with PoT weights, thus with bit shifting instead of multiplication, along with appropriate encoding of the weights.
    \item Implementation and comparison of hardware accelerator of the convolution layer based on uniform quantisation and logarithmic quantisation, in particular, demonstrating a significant reduction in power demand when using PoT weights, also for high clock frequencies.
    \item A pruning algorithm adapted to logarithmic quantisation, so that the computational complexity and thus the energy requirement can be further reduced.
\end{itemize}

The remainder of the article is organised as follows. In Section \ref{sec::pot_quant} we describe the algorithmic basis of PoT quantisation. In Section \ref{sec::prev_work}, we present a brief summary of related work. Section \ref{sec::hw_design} describes the hardware implementation of convolution filter modules, which are then used to implement a hardware accelerator, the benchmark results of which are summarised in Section \ref{subsec::benchmark_results}. The proposed pruning algorithm is described in Section \ref{subsec::pruning}. Section \ref{sec::conclusion} summarises the results of the experiments performed.
\section{Power-of-Two Quantisation} \label{sec::pot_quant}
By default (e.g. in high-end GPUs), neural network computations are implemented using at least 32-bit floating point numbers.
Quantisation involves reducing the precision of the weights, but also the activations, in such a way that they can be represented on fewer bits, in particular as integers.
The most common quantisation method is uniform quantisation, which for a given set of numbers (e.g., weights in a neural network layer) defines equidistant quantisation levels.
Logarithmic quantisation works differently, guaranteeing a higher density of quantisation levels around zero and thus mimicking well the distribution of weights in a neural network.



Logarithmic quantification $LQ$ is defined by Equation \eqref{eq::loq_quant}, where $w$ is the floating point weight and the Full Scale Range (FSR) constant is responsible for determining the minimum and maximum quantisation levels.
\begin{equation} \label{eq::loq_quant}
\begin{split}
&LQ(\mbox{w, bitwidth, FSR}) = \begin{cases} 0, & \mbox{if } w = 0 \\ 2^{\tilde{w}}, & \mbox{otherwise} \end{cases} \\ \\
&\tilde{w} =  \mbox{clip}(\mbox{round}(log_{2}(|w|)), \mbox{FSR}-2^{\mbox{bitwidth}}, \mbox{FSR}) \\ \\
&\mbox{clip(w, min, max)} = \begin{cases} 0, & w \leq min \\ max - 1, & w \geq max \\ w, & \mbox{otherwise} \end{cases}
\end{split}
\end{equation}
Following \cite{pot:DPR}, the weights of a given network layer $W$ are first normalised to the interval $[-1, 1]$: $W_N = W / SF$, where SF is a Scaling Factor equal to $SF = max(abs(W))$. 
Then, $W_N$ is transformed according to Eq. \eqref{eq::loq_quant} and the quantised weights $|W_Q| = 2^{LQ(W_N)} \times SF$ are obtained. 
The $LQ(W_N)$ are powers of two, allowing the multiplication in the convolution operation to be reduced to a bit shift, and thus greatly simplifying the computations.

Consider an example convolution filter $$W = [ 0.0034, -0.12, 0.045, 0.2, 1, -1.05, 2.34, -0.44, 0.5]$$ 
The subsequent steps for 4 bit width quantisation will be as follows:
\begin{equation*}
\begin{gathered}
\mbox{SF = max(abs(W))} = 2.34 \\
W_N = W / SF = \scriptstyle{[ 0.0015, -0.0513,  0.0192,  0.0855, 0.4274, -0.4487,  1, -0.188, 0.2137]}
\end{gathered}
\end{equation*}
The next step is to transform $W_N$ according to Eq. \eqref{eq::loq_quant} -- only the absolute values of the weights are quantised. 
We assume the value of FSR equal to 0, which corresponds to the maximum absolute value of the normalised weights ($2^0=1$). 
For the exemplar filter, we obtain:
\begin{equation*}
|W_Q| = 2^{LQ(W_N)} * SF = 2^{[-0, -4, -5, -3, -1, -1,  0, -2, -2]} * SF
\end{equation*}
The exponent vector is actually a vector of bit-shift values used for simplified multiplication with input activation. 
The direction of the shift is fixed for all weights (minus sign -- to the right), so it can be omitted.
In its place, we insert the weight sign, so the exponent vector now has the form:
\begin{equation*}
W_{EXP} = [ 0, -4,  5,  3,  1, -1,  0, -2,  2]
\end{equation*}
Thus, each weight can be stored on 4 bits: the first for the weight sign and the rest for the bit shift value. 

$SF$ is constant across the layer (the same for all filters in the layer), so it can be combined with batch normalisation gain and will not introduce additional computations load.
%
\section{Related Work} \label{sec::prev_work}
%
Weights quantisation is the basic method for reducing the computational and memory complexity of neural network.
The variety of existing approaches is well surveyed in \cite{gholami21}, with comparison between uniform and non-uniform quantization, as well as different quantisation granularity (layerwise, channelwise...), symmetry and quantisation algorithms (Quantisation Aware Training or Post Training Quantisation).

There are several papers on logarithmic quantisation in the literature -- in particular, on the quantisation algorithm itself, or Quantisation Aware Training (QAT).
Logarithmic quantisation for neural networks was first introduced in \cite{Miyashita16}. 
The paper \cite{deepshift19} proposes an approach in which the network learns the bit shift values.
The authors implemented GPU kernels for PoT-weighted computation and showed gains over standard multiplication-based filtering.
The work of \cite{APoT19} extends the PoT weights to Additive-Powers-of-Two thereby increasing the resolution of the quantisation levels. 
Although this solution represents a State-of-the-Art in terms of achieved accuracy, as shown in \cite{pot:DPR}, the need to sum two partial bit shift results, as well as the additional logic to decode the weights, introduces a higher degree of complexity compared to PoT quantisation.
This latter paper also proposes learning PoT networks using STE (Straight Through Estimator -- quantised weights are used for forward pass, while a full precision version is used for backpropagation) and demonstrates gains in using bit shifting over multiplication. 

The papers \cite{9577718} and \cite{Chmiel21} are devoted to learning algorithms, as well as \cite{Vogel18}, where the authors extend the considerations to a hardware implementation of accelerator based on logarithmic quantisation. 
The proposed PE based on bit-shifting for 5 bit weights allowed for a $22.3\%$ reduction in power consumption compared to the PE with 8 bit fixed-point multiplier design, for implementation in the Xilinx FPGA Virtex 7. 
However, it also introduced additional operations resulting from the use of a logarithmic base different from 2.
Similarly, in \cite{Xu20} the authors drop the logarithmic base 2, and show energy gains for using 5-bit logarithmic quantisation over the standard 16-bit multiplier.
Finally, the article \cite{Lee17} presents dynamic power consumption for a hardware implementation of matrix multiplication for uniformly and PoT quantised values. With 3 bit width weights, one can assume a decrease in power consumption of about $20\%$, for 5 bits -- $35\%$, in favour of PoT weights.

This paper complements the research described above by showing energy gains resulting from the use of PoT weights for a hardware accelerator implemented in FPGA.
\section{Hardware Design} \label{sec::hw_design}
\begin{figure}[!t]
\begin{center}
\includegraphics[width=0.5\textwidth]{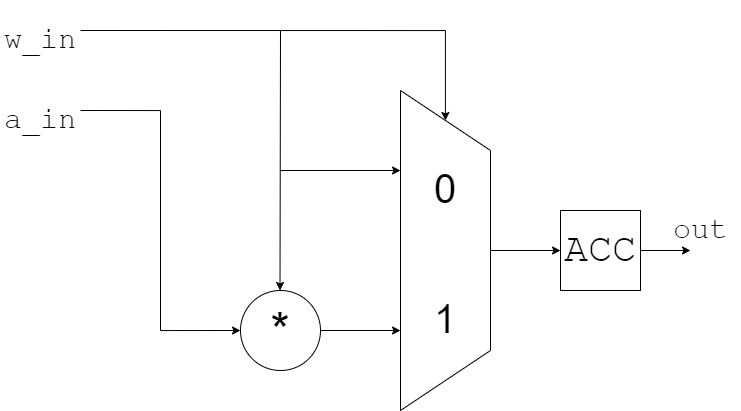}
\caption{Schematic of a simple Multiply Accumulate unit with zero weight detection.} \label{fig::mac_uniform}
\end{center}
\end{figure}

\begin{figure}[!t]
\begin{center}
\includegraphics[width=0.8\textwidth]{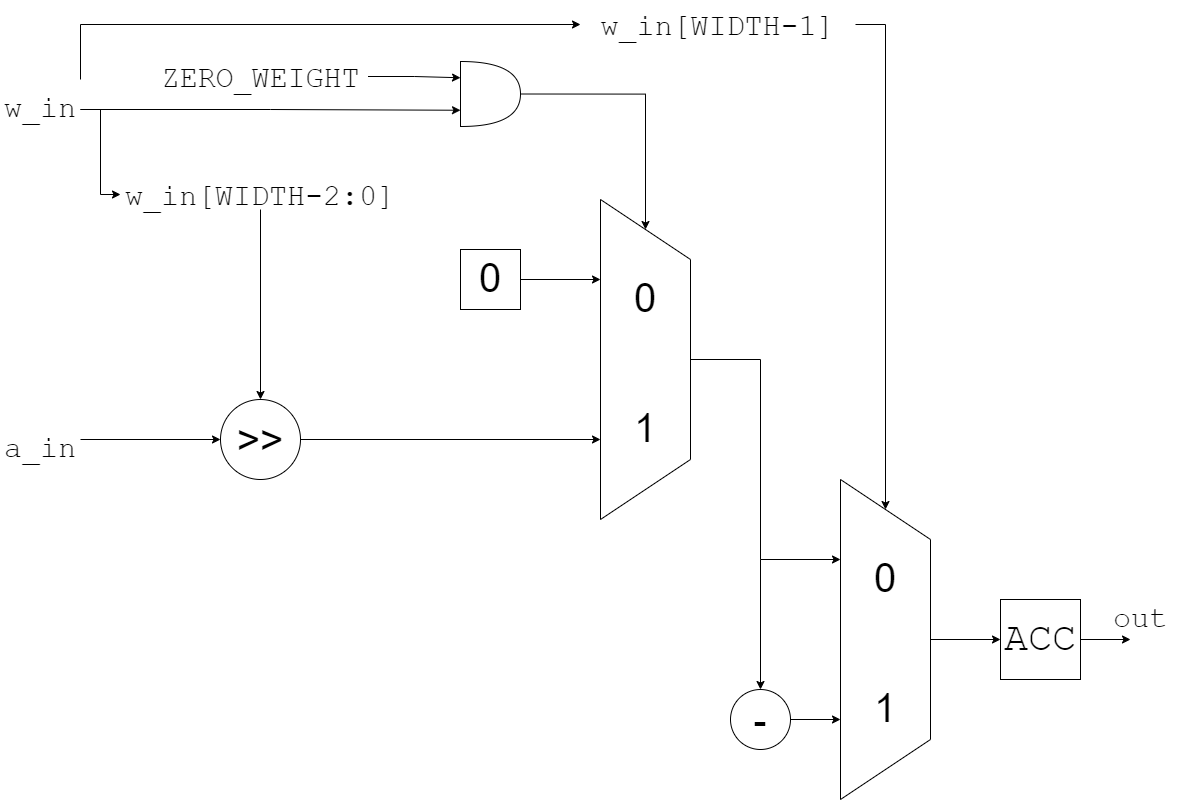}
\caption{Schematic of Bitshift Accumulate unit with zero weight detection and proper weight encoding. For bitshifting we use only the absolute value of the weight (we always do a right bit shift). After bitshifting, we use the weight sign bit to decide whether the obtained partial product should be added or subtracted from accumulator (it is mandatory since weights are not coded in two's complement).} \label{fig::mac_pot}
\end{center}
\end{figure}

%
To compare the acceleration of a convolution layer for networks with uniform and PoT quantisation, two modules were designed to perform the convolution operation -- one based on Multiply and ACcumulate (MAC -- Fig. \ref{fig::mac_uniform}) and the second on Bitshift and ACcumulate (BAC -- Fig. \ref{fig::mac_pot}) unit.
In both cases, a test has been added to see if the weight is equal to zero, which allows the multiplication (bit shift) to be possibly skipped and thus reduces the actual power consumption.
The obvious difference between the modules is the way the multiplication is performed -- simplified to a bit shift for the PoT weights. 
An additional aspect that needs to be solved is how to encode the logarithmic weights, which consist of a sign and a absolute bit shift value -- thus, they are not written in the two's complement code. 
First, it is necessary to distinguish between a zero weight and a weight with a power equal to $0$ (i.e., thus with absolute value $1$). 
According to Equation \eqref{eq::loq_quant}, for weights encoded on 4 bits, we will obtain $14$ different levels (\{6, -5, ..., -0, 0, ..., 5, 6\}), which therefore leaves the possibility of encoding the zero weight in two ways.
Second, the sign of the weight is encoded independently of the bit-shift value itself, so partial product accumulation must distinguish between the case where the weight is negative (in which case subtraction occurs) and positive (addition). 
Both modules -- MAC and BAC -- operate with the same latency, i.e., they need two clock cycles to compute the partial weighted sum of the inputs. The computation is streamlined, which means that for a filter of size $NxN$, the convolution result will be obtained after $N*N+1$ clock cycles.
Both designs have clock, enable, and reset (for accumulator reset) signals, which are used by the external module to control the data flow.
\subsection{Benchmark Results} \label{subsec::benchmark_results}
%
The proposed modules were implemented in programmable logic (FPGA) of the Zynq UltraScale + MPSoC ZCU104 platform using the Vivado 2020.2 software.
For a proper and fair comparison of the two modules, the use of DSPs (customarily used for multiplication operations) was disabled in the synthesis process, which forced the implementation of multiplication using LUTs for the MAC unit.
Finally, the Aera Explore implementation strategy was chosen, although it should be noted that for the others (Power Explore, default) the obtained results were not significantly different -- due to the low complexity of the proposed designs.
Table \ref{tab::units} shows the resource usage for single MAC and BAC modules, assuming that network weights are, regardless of method, quantised to 4 bits and activations to 8 bits.
Despite the additional logic needed to accumulate bit-shift results depending on the sign of the weight, the BAC module needs about $1.4x$ fewer LUTs. 
FFs are responsible for registers (memory cells) and the difference in their use, although slight, is also due to the change from a multiplication operation to a bit shift.
\begin{table} \label{tab::units}
\caption{Comparison of the post-implementation resources used for single MAC and BAC units with input activation x weight bit widths values.}
\begin{center}
\begin{tabular}{l@{\quad}cc}
\hline
\textbf{MAC PE}  &     \textbf{LUT} & \textbf{FF} \\
\hline
\textbf{Uniform 8x4}  & 63 & 45 \\
\textbf{PoT} 8x4 & 44 & 41\\[2pt]
\hline
\end{tabular}
\end{center}
\end{table}
%

To investigate the power requirements of neural network accelerators based on the quantisation methods analysed, a sample convolution layer (from the ResNet family of networks) module was designed, consisting of $512$ filters of size $3x3$. 
The construction of a complete accelerator is another challenge, and several approaches are encountered in the literature (as well as commercial solutions), including the implementation of a ``universal'' layer, i.e., configurable in such a way that it can perform computations for both the ``smallest'' and the ``largest'' layer.
Since most modern networks have a regular structure, it usually comes down to implementing the ``largest'' layer. 
Such an accelerator then iteratively computes the results of each layer, using memory (BRAM or DRAM -- Block or Distributed RAM) to read and write the results of intermediate layers.
This approach makes it possible to implement deep networks even in small embedded devices -- for devices with more resources, instead of a single universal layer, multiple layers can be implemented, where streamlined computations are performed, and thus reduce to number of reads and writes to and from memory. 
In view of the above, the implementation of a single layer, representative in size, will allow a demonstration of the performance of the entire accelerator.

For both layers based on uniform and PoT quantisation, a similar test design was proposed.
\begin{itemize}
    \item we implemented a layer module, consisting of $512$ units, with input clock, enable and reset signals, and input weights for each filter (unit) and input activation,
    \item we created a wrapper, with BRAM instances for weights and activation, logic controlling value reading and MAC/BAC units, as well as a clock signal generator,
    \item in order to confirm the correctness of the calculations (and also to prevent optimisation of the outputs of the unconnected modules \footnote{Vivado performs multiple optimisations during design synthesis, which remove the unused elements}), we added a built-in logic state analyser (ILA) to the layer output data. 
\end{itemize}

Table \ref{tab::layers} shows the resource usage and power consumption for both layers (excluding the other elements of the test design). 
The clock was set to $350$MHz and the dynamic power consumption was estimated using Vivado Tools, based on the implemented design. 
Using PoT weights, and thus replacing the typical multiplication for convolution operations with a bit shift, allowed for an increase in power efficiency of about $1.4x$.
At the same time, it should also be noted, as shown in the paper \cite{pot:DPR}, that the accuracy of low-precision networks (e.g., 4 bit width) is much higher using logarithmic quantisation than for those with a linear distribution.
The above also confirms the results of the synthesis of single MAC and BAC units shown in \cite{pot:DPR}.
\begin{table} \label{tab::layers}
\caption{Resources usage and dynamic power consumption (estimated using Vivado 2020.2) for MAC and BAC based layers with $512$ filters.}
\begin{center}
\begin{tabular}{l@{\quad}cccc}
\hline
\textbf{MAC PE}  &     \textbf{LUT} & \textbf{FF} & \textbf{Clock [MHz]} & \textbf{Power [W]} \\
\hline
\textbf{Uniform 8x4} & 24193 & 23265 & 350 & $0.303$ \\
%
\textbf{PoT} 8x4 & 19557 & 17024 & 350 & $0.214$ \\[2pt]
\hline
\end{tabular}
\end{center}
\end{table}
%

\begin{figure}[!t]
\begin{center}
\includegraphics[width=0.7\textwidth]{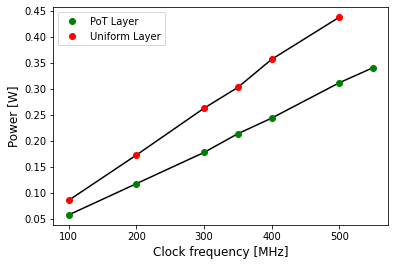}
\caption{Relation between the clock frequency and power consumption for both the uniform and the PoT layer: for higher frequencies the gap between the MAC and the BAC layer is more visible, with the BAC layer consuming around $0.6$ of the power needed by the MAC layer.} \label{fig::clock_vs_power}
\end{center}
\end{figure}

The actual processing time of the input frame depends on the latency of the modules (identical for both MAC and BAC units), as well as the clock frequency. 
To investigate the approximate limit of this frequency, we conducted a series of experiments summarised in Fig. \ref{fig::clock_vs_power}, showing the relationship between the clock rate and the dynamic power consumed by the convolution layer module.
For the MAC layer, the highest operating frequency is $500$ MHz, for the higher value of $550$MHz the implementation failed to meet the timing constraints.
For the BAC layer, a similar problem occurred only at $600$ MHz.
%
As described in Section \ref{sec::prev_work}, most of the existing works show decreases in the power used relative to standard (e.g. 16 bits) multipliers.
For low-precision computation processing elements, the work of \cite{Lee17} indicated a decrease of about $20\%$, with 5 bits -- $35\%$, in favour of PoT weights over uniformly quantised ones.
Our implementation allows for a $40\%$ reduction in dynamic power at 4 bit width weights.
Because the aforementioned publication is sparing in its description of the design, we are unable to make a more accurate comparison.

\subsection{Going Lower on Power - Pruning} \label{subsec::pruning}
%
A further reduction in energy consumption can be achieved by introducing more zero weights using pruning.
Uniform quantisation for weights with very low precision (e.g., 4 bit width) introduces automatic pruning that does not affect the number of quantisation steps in any way -- values below the minimum quantised weight (different from zero) are rounded to zero, resulting in a relatively high percentage of zero weights.
This is different for logarithmic quantisation: due to the dense distribution of quantisation levels around zero, the number of quantisation levels is reduced by discarding the lowest ones, when weights with low absolute values are zeroed out.
This can significantly affect the efficiency of the network at very low computational precision.

At the same time, properly performed pruning can result in the zeroing of up to a few tens of percent of weights having little effect on the final network indication, which in turn can result in further gains (after introducing a bit shift in place of multiplication) gains in hardware implementation:
\begin{itemize}
    \item In \cite{pot:DPR}, authors show that zeroing out the weights allows compression and thus reduces the memory requirements (which also implies a decrease in energy requirements).
    \item The designed MAC and BAC units skip the multiplication/bit shifting operations when zero weight is detected. Depending on the network architecture and the degree of pruning, this will result in smaller or larger energy gains during system operation.
\end{itemize}

\begin{figure}[!t]
\begin{center}
\includegraphics[width=\textwidth]{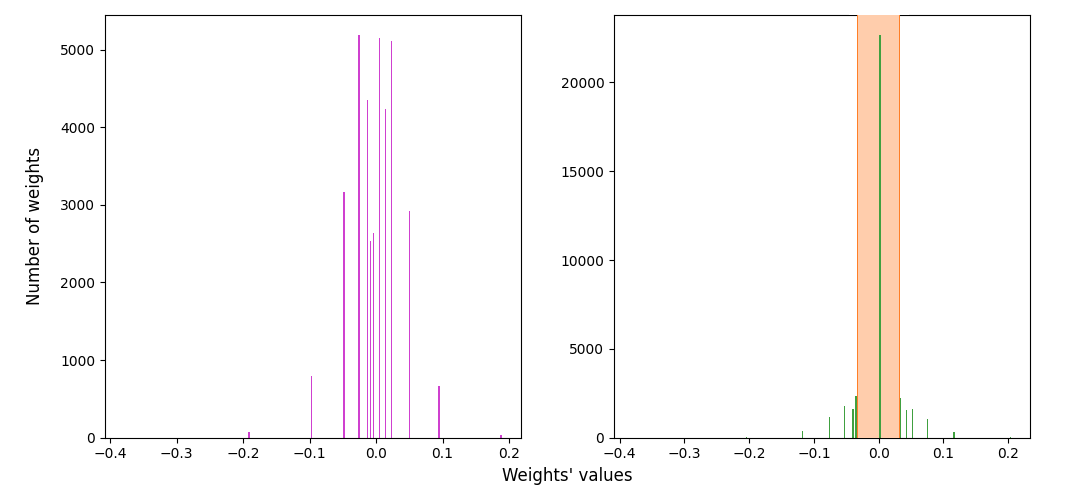}
\caption{Visualisation of the proposed pruning method: the left plot shows the distribution of 4-bit width PoT quantised weights of an exemplar layer of ResNet18 trained on ImageNet. After applying pruning, the minimum quantisation value is shifted away from zero, creating a dead zone -- marked in red in the right plot.} \label{fig::pruning_deadzone}
\end{center}
\end{figure}

In view of the above, a pruning algorithm adapted to the peculiarities of logarithmic quantisation is proposed, introducing the so-called \textit{dead zone} around zero (in the distribution of weights of a given layer) by the following modification of the algorithm described in Section \ref{sec::pot_quant}:
\begin{enumerate}
    \item After normalising the absolute values of the weights to the $[0, 1]$ interval (and getting $W_N$) we implement pruning ($W_P$) with a fixed degree, e.g. relative to the weight with the maximum modulus (i.e. 1).
    \item We normalise the pruned weights, excluding zero weights, to the interval $[0, 1]$.
    \item We perform logarithmic quantisation.
\end{enumerate}
Thus, the value of the Scaling Factor is equal to $SF = w\_max - w\_min$, where $w\_min$ is different from zero. To reconstruct the proper quantised weight value, we must then introduce an offset equal to $OFFSET = w_min$, so now $W_Q=Q(W_P) * SF + OFFSET$.
Thus, we move the minimum quantisation level away from zero, creating a large enough 'dead zone' while keeping the logarithmic distribution outside of it -- Fig. \ref{fig::pruning_deadzone}.
%

To evaluate the proposed pruning method, ResNet20 was trained on the CIFAR10 dataset, with varying degrees of pruning -- weights below a value set of $PF*max\_weight$ were set to zero. 
The STE algorithm is used to train the quantised, PoT 4 bit width network.
That is, we initialise the weights with the full precision version, and then in each training iteration, we do forward pass with the quantised weights and backward pass using floating point values.
\begin{table} \label{tab::pruning}
\caption{Pruning experiments for the 4-bit width PoT quantised ResNet20 trained on CIFAR10. Increasing the value of Pruning Factor causes more weights to be set to zero, while retaining very good accuracy of the 4-bit width quantised network.}
\begin{center}
\begin{tabular}{l@{\quad}cc}
\hline
\textbf{PF}  &     \textbf{Accuracy} & \textbf{Number of zeros} \\
\hline
no pruning  & $ 91.11\% $ & $ 0 $ \\
$0.05$ & $ 91.62\% $ & $ 21\% $\\
$0.1$ & $ 91.54\% $ & $ 41\% $\\
$0.2$ & $ 90.43\% $ & $ 70\% $\\[2pt]
\hline
\end{tabular}
\end{center}
\end{table}
%
Table \ref{tab::pruning} shows the results of the experiment. 
Since the tested network is very redundant, and due to the introduction of a 'dead zone' that allowed all quantisation levels to be maintained, the decrease in network performance is small even when $70\%$ of weights are set to zero. 
We also observed a slight increase in performance for lower PF values, indicating that the introduced pruning prevented overfitting.

The presented pruning algorithm, in combination with the designed BAC module presented in Section \ref{sec::hw_design}, which allows one to skip the computation for zero weights, will therefore allow one to further significantly reduce the power requirements of the real hardware accelerator for sparse neural networks.
\section{Conclusion} \label{sec::conclusion}
%
In this article, we demonstrate the effect of logarithmic quantisation on energy demand in an FPGA-based accelerator of convolutional neural networks.
The replacement of the multiplication operation with a bit shift reduced the power requirement by $1.4x$.
Moreover, a pruning algorithm adapted to logarithmic quantisation is proposed, so that while maintaining high network efficiency, power consumption can be further reduced.
In view of this and previous works, it is important to conclude that neural network quantisation using PoT weights, compared to uniform quantisation, allows for highly energy-efficient acceleration in embedded devices.
Further work should focus on the hardware implementation of a complete vision system based on the presented solution.
The use of PoT weights and the proposed convolution filter hardware design shall allow for real-time and energy-efficient implementation of particularly complex, state-of-the-art deep convolutional neural networks.

\subsubsection{Acknowledgements} The work presented in this paper was supported by the AGH University of Science and Technology project no. 16.16.120.773.

%
%

\bibliographystyle{bibtex/splncs03_fixed}
\bibliography{DPR-TK-ICCVG22}

\end{document}